\documentclass[9pt,conference]{IEEEtran}
\IEEEoverridecommandlockouts

\usepackage{algorithm}
\usepackage{multirow}
\usepackage[inkscapearea=page]{svg}
\usepackage{subfigure}

\usepackage{hyperref}
\usepackage{pifont}
\usepackage{multirow}
\usepackage{float}
\usepackage{xspace}

\usepackage{cjhebrew}
\usepackage{booktabs}
\usepackage{color}
\usepackage{etoolbox}
\usepackage{amsmath}

\usepackage{cite}
\usepackage{amsmath,amssymb,amsfonts}
\usepackage{algorithmic}
\usepackage{graphicx}
\usepackage{textcomp}
\usepackage{xcolor}
\def\BibTeX{{\rm B\kern-.05em{\sc i\kern-.025em b}\kern-.08em
    T\kern-.1667em\lower.7ex\hbox{E}\kern-.125emX}}

\usepackage{etoolbox}
\newtoggle{release}
\usepackage{cleveref}

\newcommand{\newpara}[1]{\vspace{0.075cm} \noindent {\bf #1}}

\togglefalse{release}

\iftoggle{release}{
\newcommand{\adios}[1]{}
\newcommand{\oa}[1]{}
\newcommand{\av}[1]{}
}
{
\newcommand{\adios}[1]{\textcolor{red}{YA: #1}}
\newcommand{\oa}[1]{\textcolor{blue}{OA: #1}}
\newcommand{\av}[1]{\textcolor{green}{av: #1}}
}

\newcommand{\pmr}[1]{\scriptsize$\pm$#1}

\newcommand{\pmt}[2]{$#1\pm#2$}

\title{Unsupervised Speech Segmentation: \\ A General Approach Using Speech Language Models}
\author{\IEEEauthorblockN{Avishai Elmakies}
\IEEEauthorblockA{\textit{Hebrew University of Jerusalem}}
\and
\IEEEauthorblockN{Omri Abend}
\IEEEauthorblockA{\textit{Hebrew University of Jerusalem}}
\and
\IEEEauthorblockN{Yossi Adi}
\IEEEauthorblockA{\textit{Hebrew University of Jerusalem}} \\ 
}

\begin{document}

\maketitle

\begin{abstract}
    In this paper, we introduce an unsupervised approach for Speech Segmentation, which builds on previously researched approaches, e.g., Speaker Diarization, while being applicable to an inclusive set of acoustic-semantic distinctions, paving a path towards a general Unsupervised Speech Segmentation approach. Unlike traditional speech and audio segmentation, which mainly focuses on spectral changes in the input signal, e.g., phone segmentation, our approach tries to segment the spoken utterance into chunks with differing acoustic-semantic styles, focusing on acoustic-semantic information that does not translate well into text, e.g., emotion or speaker. While most Speech Segmentation tasks only handle one style change, e.g., emotion diarization, our approach tries to handle multiple acoustic-semantic style changes. Leveraging recent advances in Speech Language Models (SLMs), we propose a simple unsupervised method to segment a given speech utterance. We empirically demonstrate the effectiveness of the proposed approach by considering several setups. Results suggest that the proposed method is superior to the evaluated baselines on boundary detection, segment purity, and over-segmentation. Code is available at \url{https://github.com/avishaiElmakies/unsupervised_speech_segmentation_using_slm}. 
\end{abstract}
\begin{IEEEkeywords}
Speech Language Models, Speech Segmentation.
\end{IEEEkeywords}

\section{Introduction}
\label{section:Intro}
\emph{Speech segmentation} is the task of breaking a speech signal into individual sound units. It plays an essential role in a variety of speech and audio applications such as Automatic Speech Recognition~\cite{kubala1996transcribing, rybach2009audio}, speaker diarization~\cite{moattar2012review}, and speech science~\cite{adi2016vowel,adi2016automatic}. Most prior work on speech segmentation focused on spectral changes in the spoken utterance, as a result, it mainly captures phone or phoneme segments~\cite{rybach2009audio, kamper2022word, adi2017sequence, kreuk2020self, kreuk2020phoneme, cuervo2022variable}. Although much progress has been made with spectral segmentation approaches, most of these methods focus on a single acoustic-semantic style change, e.g., emotion diarization ~\cite{wang2023speechemotiondiarizationemotion}, usually using supervised methods. 

\emph{Speech Language Models (SLMs)} are a promising research direction in the field of speech and audio processing~\cite{lakhotia2021generative,borsos2023audiolm, kreuk2022textless, hassid2024textually}. SLMs first represent speech and audio signals as discrete acoustic units, on which a language model is applied to maximize the sequence likelihood~\cite{lakhotia2021generative, kharitonov2022textless}. This method was shown to be beneficial in several speech modeling and generation tasks~\cite{lakhotia2021generative,borsos2023audiolm, hassid2024textually}. Improving the semantic and acoustic abilities of SLMs is an ongoing topic of research in the field ~\cite{kharitonov2022text}. The ability to directly obtain good estimates of the distributions of speech signals without converting them into text allows us to use SLMs for applications that cascading methods are not well equipped to handle, e.g., emotion-related tasks. 

\begin{figure}[t!]
    \centering
    \includegraphics[width=0.9\columnwidth]{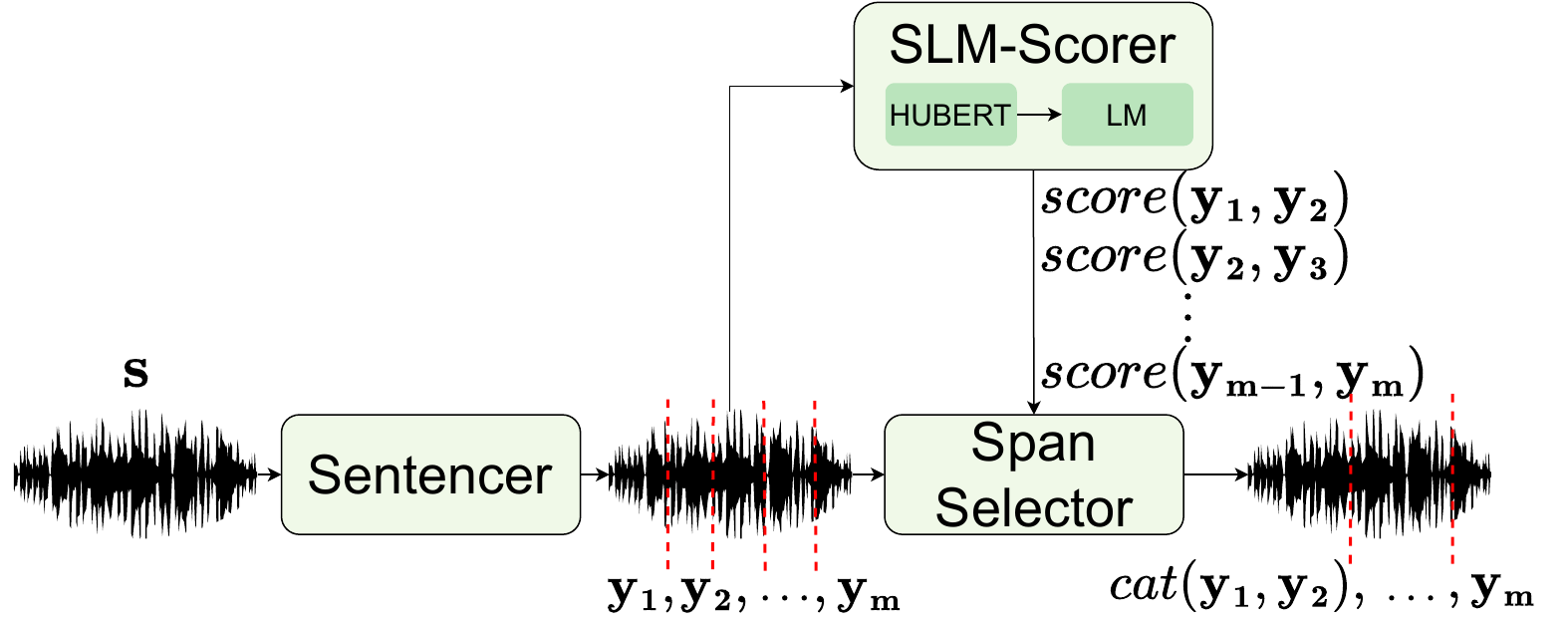}
    \caption{The proposed pipeline. Speech utterance is first segmented into $m$ segments (``acoustic-sentences'') using the sentencer, the SLM scores consecutive segments and the span selector selects the ones to merge.\label{figure:pipeline}}
    \label{fig:example}
\end{figure}

\emph{In this work}, we define an unsupervised approach for speech segmentation, focusing on acoustic-semantic style changes that do not translate well into text. We leverage the progress in the field of SLMs and show a simple pipeline that tackles this new approach for unsupervised speech segmentation. Specifically, we first break the audio into segments (``acoustic-sentences''). Then, we score consecutive sentences using the probabilities obtained from the SLMs. Finally, we select acoustic-sentences to merge using the calculated scores. A visual description of the proposed method can be seen in Fig.~\ref{figure:pipeline}. We examine the proposed approach considering two case studies of our proposed approach: (i) emotions; and (ii) gender. We conduct evaluation under controlled settings using synthetically generated datasets. Results suggest that although simple, the proposed pipeline is found to be highly efficient, providing superior performance to the evaluated baselines considering boundary detection, over-segmentation, and segment purity. 

\newpara{Our contribution:} We introduce a general approach for Unsupervised Speech Segmentation, which, unlike previous work, focuses on acoustic-semantic style concepts rather than spectral changes. We provide a simple and effective unsupervised method based on SLMs. We empirically demonstrate the efficiency of the proposed approach considering the segmentation of several acoustic-semantic concepts.

\section{Background}

In speech segmentation, we get a speech signal $\mathbf{s} \in \mathbb{R}^T$, where $T$ is the length of the sequence, and divide it into $m$ segments $\mathbf{y_1},\dots,\mathbf{y_m}$, where the length varies between each $\mathbf{y_i}$. In our setup, we aim to segment the speech utterance into acoustic-semantic style segments. As we consider the unsupervised setup, the acoustic-semantic style segments can have different forms, e.g., a change of emotion, a change of gender, a change of speaker, etc. 

Language models (LMs) have been a topic of recent discussion in various fields of AI, ranging from NLP~\cite{wei2023overview} to visual LMs~\cite{esser2021taming}. Recently, Speech Language Models (SLMs) also emerged as an interesting line of research~\cite{lakhotia2021generative, kharitonov2022text, borsos2023audiolm}. 

SLMs operate similarly to textual LMs using next-token prediction. However, unlike text, speech signals are continuous in nature; hence, it is not clear how to represent speech in a discrete manner. One common approach to mitigate this, is to quantize contextualized speech representation obtained from a pre-trained self-supervised model~\cite{lakhotia2021generative, borsos2023audiolm, hassid2024textually}. This is often done using the k-means algorithm. The tokens generated following the aforementioned approach have been shown to correlate with phonemes~\cite{Sicherman_2023}. Next, such speech tokens can be directly used for the task of next-token prediction in a similar way to textual LMs. Using both a discretization model and a language model allows us to approximate the distribution of our data, for an audio signal $\mathbf{s}$, using some discretization model $f_{\phi}$, such that $\phi$ are the parameters of the discretization model, we get the following:
\begin{equation}
    P_{real}(\mathbf{s}) \approx P_\theta(f_\phi(\mathbf{s})),f_\theta(\mathbf{s}) = (x_1,\dots, x_n) \label{equation:slm_approx}
\end{equation}
where $(x_1,\dots, x_n)$ are the speech tokens fed into the SLM $P_\theta$. In this paper, we show how to use this approximation for unsupervised segmentation of speech utterances.

\section{Method}\label{section:ours}
In this section, we introduce a simple and flexible pipeline for tackling Unsupervised Speech Segmentation. This pipeline is inspired by text-based segmentation methods, which work on sentences and scores to segment the text~\cite{wagner-etal-2022-topical}. The pipeline can be seen in Fig.~\ref{figure:pipeline} and is composed of 3 parts: (a) a sentencer that splits the audio into segments (``acoustic-sentences''), (b) a scorer that scores consecutive sentences, and (c) a span-selector that uses the scores and the sentences to select the final spans/segments. 

\subsection{Sentencer}
Similarly to~\cite{wagner-etal-2022-topical} we use a naive and straightforward approach to perform the initial segmentation. We segment the audio into equally sized segments, which we denote as \emph{acoustic-sentences}. This process can be considered as an initial guess for the segments, where in later stages we will refine such segmentation. 

\begin{table*}[t!]
  \caption{Results for C(10), A(10) and T(-10) described in Section~\ref{subsection:spanselctor}, comparing Equal Length (EL), Contrastive Learning Score (CLS), Diarization Methods (DM), and PMI model (ours), and comparing the threshold span selector with the best threshold value to the other approaches. Confidence intervals use $\alpha = 0.9$. \label{tab:mainres}}
  \centering
  \resizebox{\textwidth}{!}{ 
  \begin{tabular}{ l|c|c|c|c|c|c|c|c|c|c|c|c }
    \toprule
    \multirow{3}{*}{} & \multicolumn{6}{c|}{\textbf{Change of emotion}} & \multicolumn{6}{c}{\textbf{Change of gender}}  \\
    \cline{2-13}
    & \multicolumn{3}{c|}{\textbf{EMOV-DB}} & \multicolumn{3}{c|}{\textbf{IEMOCAP}} & \multicolumn{3}{c|}{\textbf{EMOV-DB}} & \multicolumn{3}{c}{\textbf{IEMOCAP}} \\ 
    \cline{2-13}
    & \textbf{PR-F1} & \textbf{R-Val} & \textbf{PC-F1} & \textbf{PR-F1} & \textbf{R-Val} & \textbf{PC-F1} & \textbf{PR-F1}  & \textbf{R-Val} & \textbf{PC-F1} & \textbf{PR-F1} & \textbf{R-Val} & \textbf{PC-F1} \\
    \midrule
    & \multicolumn{12}{c}{C(10)}\\
    \midrule
    EL & 16.9\pmr{0.5}& 25.4\pmr{1.1}& \textbf{65.9\pmr{0.3}}& 17.2\pmr{0.4}& 25.3\pmr{1.0}& \textbf{67.5\pmr{0.2}}& 10.8\pmr{0.3}& 33.1\pmr{0.2}& \textbf{52.4\pmr{0.5}} & 10.6\pmr{0.3}& 33.0\pmr{0.2}& \textbf{59.7\pmr{0.3}}\\
    CLS       & 23.5\pmr{0.4}& 29.6\pmr{1.0}& 50.0\pmr{0.6}& 24.2\pmr{0.4}& 30.0\pmr{0.9}& 64.6\pmr{0.3}& 18.5\pmr{0.4}& 36.9\pmr{0.2}& 38.0\pmr{0.5}& 17.9\pmr{0.3}& 36.7\pmr{0.2}& 52.9\pmr{0.4}\\
    PMI     & \textbf{28.2\pmr{0.4}}& \textbf{32.3\pmr{1.1}}& 62.4\pmr{0.4}& \textbf{27.3\pmr{0.4}}& \textbf{31.7\pmr{0.9}}& 66.9\pmr{0.3}& \textbf{21.8\pmr{0.4}}& \textbf{38.7\pmr{0.2}}& 47.5\pmr{0.5}& \textbf{20.7\pmr{0.4}}& \textbf{38.2\pmr{0.2}}& 55.6\pmr{0.4}\\
    \midrule
    & \multicolumn{12}{c}{A(10)}\\    
    \midrule
    EL & 23.3\pmr{0.5}& 30.8\pmr{0.5}& \textbf{74.9\pmr{0.2}}& 22.5\pmr{0.4}& 28.7\pmr{0.5}& 70.1\pmr{0.2}& 22.4\pmr{0.4}& 31.8\pmr{0.4}& \textbf{74.3\pmr{0.1}}& 21.7\pmr{0.3}& 29.2\pmr{0.4}& 69.8\pmr{0.1}\\
    PMI    & \textbf{33.2\pmr{0.4}}& \textbf{41.4\pmr{0.4}}& 72.0\pmr{0.2}& \textbf{31.9\pmr{0.4}}& \textbf{39.0\pmr{0.5}}& \textbf{73.0\pmr{0.2}}& \textbf{33.1\pmr{0.4}}& \textbf{42.4\pmr{0.4}}& 69.8\pmr{0.2}& \textbf{33.2\pmr{0.3}}& \textbf{41.2\pmr{0.3}}& \textbf{72.6\pmr{0.1}}\\
        \midrule
    & \multicolumn{12}{c}{DM}\\
        \midrule

    SD & 40.7\pmr{0.8}& 44.5\pmr{0.8}& 75.2\pmr{0.2}& 33.1\pmr{0.4}& 34.8\pmr{0.6}& 74.0\pmr{0.2}& 43.9\pmr{0.6}& 47.3\pmr{0.7}& 75.6\pmr{0.1}& 42.9\pmr{0.6}& 37.8\pmr{0.7}& 75.3\pmr{0.2} \\
    ED & 16.7\pmr{0.6}& -95.6\pmr{9.4}& 71.5\pmr{0.3}& 24.6\pmr{0.4}& -187.9\pmr{5.6}& 77.7\pmr{0.2}& 9.6\pmr{0.4}& -68.7\pmr{8.8}& 40.9\pmr{1.0}& 16.6\pmr{0.4}& -250.0\pmr{13.7}& 73.2\pmr{0.4}\\
    \midrule
    & \multicolumn{12}{c}{T(-10)}\\
    \midrule   
    PMI    & 30.3\pmr{0.5}& 41.2\pmr{0.4}& 65.9\pmr{0.4}& 32.2\pmr{0.4}& 22.4\pmr{1.0}& 73.0\pmr{0.2}& 30.5\pmr{0.4}& 43.7\pmr{0.3}& 63.4\pmr{0.3}& 32.4\pmr{0.3}& 11.5\pmr{1.8}& 70.8\pmr{0.2}\\
     \bottomrule
  \end{tabular}}
\end{table*}

\subsection{Scorer}
\label{subsection:Scorer}
Following~\cite{wagner-etal-2022-topical} we use the Point-Wise Mutual Information (PMI) which is defined for two sentences $x,y$ as follows:
\begin{equation}
    PMI(x,y) = \log \Big(\frac{P(x,y)}{P(x)\cdot P(y)}\Big). \label{equation:pmi}
\end{equation}

We use this score since PMI is a measure of association, it compares the probability of $x, y$ showing one after the other to what this probability would be if $x$ and $y$ were independent. If the PMI is small or negative, it suggests that $x$ and $y$ are more likely to be independent, making the placement of a boundary between them a reasonable hypothesis. PMI has also been shown to work well for segmentation and boundary detection in other fields \cite{wagner-etal-2022-topical,isola2014crisp}.

As we have seen, using SLMs we can approximate $P_{real}$ of audio, hence we can use both ~\eqref{equation:slm_approx} and ~\eqref{equation:pmi} to define the following score:
\begin{equation}
    score(\mathbf{y_i},\mathbf{y_{i+1}}) = \log \Big(\frac{P_\theta(f_\phi(\mathbf{y_i},\mathbf{y_{i+1}}))}{P_\theta(f_\phi(\mathbf{y_i}))\cdot P_\theta(f_\phi(\mathbf{y_{i+1}}))}\Big). 
\end{equation}

Later, this score will be used to select the number of segments and what \emph{acoustic-sentences} should be merged to reach the final segmentation of the speech utterance.

\subsection{Span selector}
\label{subsection:spanselctor}
We present three methods for selecting both the number of segments (denoted by $k$) and the segments themselves using the scores obtained from the Scorer.

\newpara{Constant number of segments.} Under this method, we first set the value of $k$. Then, boundaries are placed on the $k-1$ smallest scores to yield $k$ segments. We denote this method as C($k$). We explored values of $k \in \{10,15,20\}$.

\newpara{Adaptive number of segments.} Here, the number of segments is based on the number of initial segments (i.e., acoustic-sentences) in the speech utterance. Meaning that each speech signal will have a varying amount of segments. The overall number of segments is defined as follows, 
\begin{equation}
    k = \frac{max(0, m - 20)}{v} + 4,
    \label{equation:adaptive}
\end{equation}
where $m$ is the number of acoustic-sentences in a given signal and $v$ represents the number of acoustic-sentences needed to increase the number of segments by $1$. We subtract $20$ from the number of acoustic-sentences to prevent over-segmentation for short files and include $4$ as the minimum number of segments. We select those numbers assuming less than $10$ seconds of speech (with acoustic-sentences of $0.5$ seconds) is short for semantic segmentation of more than $4$ segments. Finally, following ~\eqref{equation:adaptive} we get $k$ and choose the boundaries as seen previously. We denote this method as A($v$). We experimented with $v\in \{5,10,15,20\}$. 

\newpara{Threshold.} The last span selection approach also selects the number of segments, $k$, dynamically. First, a threshold $t$ is used, then all scores smaller than $t$ are selected as boundaries. We denote this method with T($t$). We experimented with $t\in \{-5,-8,-10,-12.5,-15\}$.

\section{Datasets}\label{section:Datasets}

We use two benchmarks to create synthetic datasets, which will be used to evaluate the proposed method: (i) The \emph{Emotional Voices Database (EmoV-DB)}~\cite{adigwe2018emotional} data set contains English speech recordings. The recordings were obtained from four speakers (two male and two female) and contains five different emotions for most of the speakers (i.e. Neutral, Amused, Angry, Sleepy, and Disgust); (ii) The \emph{Interactive Emotional Dyadic Motion Capture (IEMOCAP)~\cite{busso2008iemocap}} dataset which contains $12$ hours of English audio-visual data (improvised and scripted) from ten speakers ($5$ men, $5$ women). We use the emotions of happy (we converted excited to happy as they are highly similar and hard to differentiate), sad, angry and neutral. 

Given these two benchmarks, we concatenate recordings from the same or different emotions to simulate acoustic-semantic style changes. We focus on the aspects of emotional style and gender style. 

\newpara{Experiment 1: change of emotion.}
In the first experiment, we focus on the change of emotion as our acoustic-semantic style change. For this, we extract the relevant utterances for each speaker and concatenate them. The number of segments is randomly chosen between $4$ and $30$, where the segments for each file are randomly chosen. All speech signals are resampled to $16$kHz whenever needed. For EmoV-DB we generated $2000$ files ($500$  files for each speaker) while for IEMOCAP 2500 we generated 2500 files ($250$ per speaker).

\newpara{Experiment 2: change of gender.} 
For this experiment, we focus on the change of gender. For each speaker, we group their utterances based on the utterances emotions. In IEMOCAP we select speakers based on the sessions in the dataset(each session contains one male and one female), while in EmoV-DB we select both speakers of the opposite gender equally. Later on, we select a random emotion. From the group of utterances for the emotion we select random segments for both speakers (in the range between $4$ to $30$). The files for both speakers are resampled to $16$kHz whenever needed and concatenated in an alternating order. We verify that there are an equal number of files that start with male and female. For EmoV-DB we generated $2000$ files ($250$ for each combination) while for IEMOCAP we generated $2500$ files ($500$ for each pair).

\section{Experimental setup}

\begin{figure*}[t!]
    \centering
    \subfigure[Metrics as a function of $k$ for span selector C($k$)]{\includegraphics[width=0.24\textwidth]{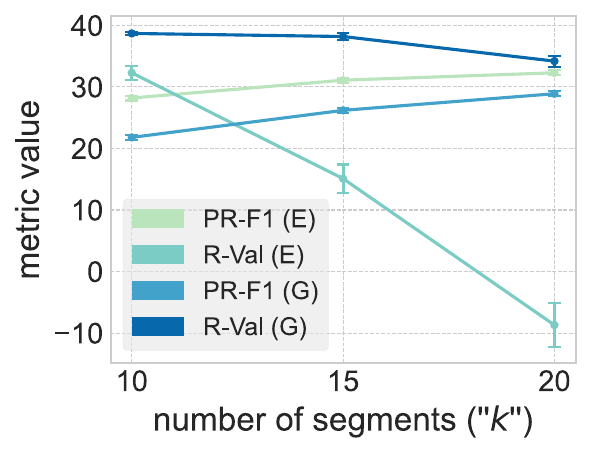}\label{subfigure:C}}\hspace{3.5pt}
    \subfigure[Metrics as a function of $v$ for span selector A($v$)]{\includegraphics[width=0.24\textwidth]{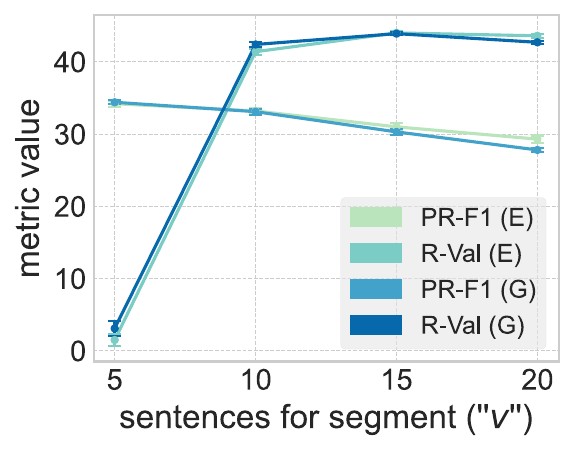}\label{subfigure:v}}\hspace{3.5pt}
    \subfigure[Metrics as a function of $t$ for span selector T($t$)]{\includegraphics[width=0.24\textwidth]{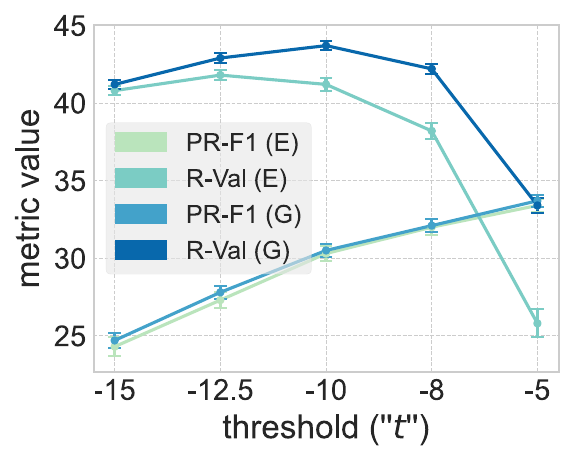}\label{subfigure:T}}\hspace{3.5pt}
    \subfigure[Metrics as a function of acoustic sentence size]{\includegraphics[width=0.24\textwidth]{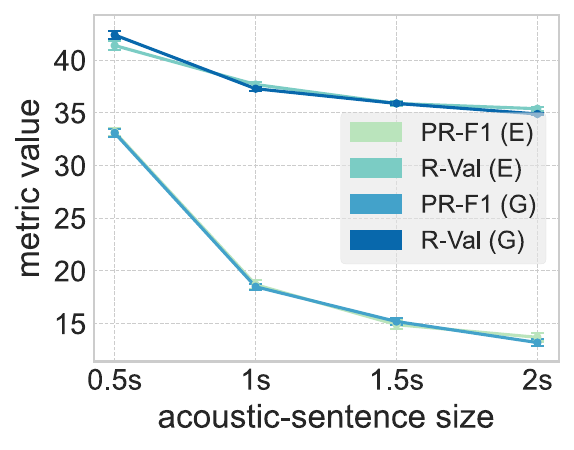}\label{subfigure:s}}
    \caption{Ablation results. Subfigures present the effect of different ablations on PR-F1 and R-Val. All results use the 350M model and are shown using EmoV-DB (E represents the ``change of emotion" experiment and G represents the ``change of gender" experiment) PC-F1 has a high positive correlation (0.55) with PR-F1 and is therefore omitted for brevity.}
    \label{figure:abaltion}
\end{figure*}

We leverage the pre-trained TWIST SLM model as described in~ \cite{hassid2024textually}, which were optimized over $\sim200$k hours of spoken data. All results in this paper are reported using the 350M parameter version. We experimented with larger models (i.e., $1.3$B and $7$B), however we did not observe significant improvements. The spoken data was first discretized using a k-means quantizer optimized over representations extracted from a pre-trained HuBERT model~\cite{hassid2024textually}. We evaluate the span selector methods and compare the results to the methods described below. We set the initial acoustic-sentences length to be $0.5$ seconds. We explored different lengths of acoustic-sentences and found that $0.5$ performs the best (see Sec~\ref{sec:abl}).

\subsection{Baselines}\label{subsection:baselines}
We evaluate the proposed method against the following baselines: 

\newpara{Equal Length (EL) Segmentor.} The first baseline is the simplest one, segmenting the audio into equal-length segments. We test this approach using both a constant number of segments and an adaptive approach to the number of segments.

\newpara{Contrastive Learning Scorer (CLS).} Next, we examine the effect of a more advanced method. We explore the usage of a scoring method as described in \ref{subsection:Scorer}. For that, we leverage a self-supervised phoneme segmentation model based on Constractive Learning to score frames. Specifically, we consider the model proposed by~\cite{kreuk2020selfsupervised}. Scoring is done via cosine similarity between adjacent frames as suggested by~\cite{kreuk2020selfsupervised}. 

\newpara{Diarization Methods (DM)}
We additionally compare the proposed approach to SOTA supervised methods used for a single acoustic-semantic style change: (i) \emph{Speaker Diarization (SD)} ~\cite{Bredin23}; (ii) \emph{Emotion Diarization (ED)} ~\cite{wang2023speechemotiondiarizationemotion}. While ED needed no changes since it gives segmentation as defined above, SD needed hyper-parameters calibration to better match our setup (i.e., SD often introduces small and unnatural segments during speaker activity which causes over-segmentation). For that, we removed segments shorter than 0.25s and combined consecutive segments that were predicted to have the same speaker with a distance of less than 0.5s.

\subsection{Evaluation}
We evaluate the proposed method considering three different evaluation metrics: (i) Recall Precision F1 (RP-F1), where we compute the recall and precision of boundary selection of the methods and compute their F1 score. For this metric we consider a tolerance of 0.5 seconds; (ii) R-Value, as proposed by~\cite{rasanen2009improved, kreuk2020selfsupervised}. We additionally consider the R-Value score to mitigate the sensitivity of the F1 score to over-segmentation; (iii) Purity Coverage F1 (PC-F1). As suggested by Bredin et al.~\cite{pyannote.metrics}, similarly to clustering, one can use segment-wise purity and coverage to test our segmentation. We include these metrics for completeness. It is important to note that only the ED model was trained on the EMOV-DB and IEMOCAP datasets. 
\section{Results}
 Results are summarized in Table~\ref{tab:mainres}. Results suggest that the proposed approach is superior to the evaluated baselines considering both PR-F1 and R-Val, meaning the proposed approach is better at finding segmentation boundaries while not performing over-segmentation. As for PC-F1, the proposed approach achieves inferior performance to the EL baseline while outperforming the CLS method. We hypothesize that this is due to the inductive bias in the datasets generation process, i.e., the length of the files has low variance. To test this hypothesis, we generated another version of the data using EmoV-DB, using only short and long audio files ($2\sigma$ from the mean). As expected, under this setup, EL segmentor A(10) performs worse than the proposed method. It achieves \pmt{69.4}{0.13} while PMI 350M A(10) reaches \pmt{74.4}{0.2} on the PC-F1 metric, in addition to being superior in the other metrics by $\sim$10 points.

 When comparing the span selection mechanism, the results suggest the adaptive approach outperforms the constant number of segments approach under all evaluated setups both in terms of PMI and EL. Interestingly, the threshold-based approach (PMI-T) achieves comparable results to both the adaptive and constant approaches on the EmoV benchmark, while producing over-segmentation on the IEMOCAP (significantly worse R-Val). This demonstrates the sensitivity of threshold calibration, as can also be seen in Fig~\ref{subfigure:T}.

 When comparing our method to SOTA ED, we see that ED performs much worse than our pipeline in all tasks, even in the change of emotion task. The model also suffers from a problem of over-segmentation. The results probably arise because of the ED's training regime. The ED model is trained only on the transition from neutral to emotion or from emotion to neutral ~\cite{wang2023speechemotiondiarizationemotion}, making our experiments an OOD test set for the model since we can have any combination of emotions one after the other. Finally, we can see that the ED model performs much better in the change of emotion experiment when compared to the change of gender experiment. Interestingly, the proposed method is more robust to the different acoustic-semantic style to segment (i.e., gender or emotion).

When comparing our method to SOTA SD, we see that our pipeline is worse in most metrics, even in the change of emotion task, which has only one speaker. We hypothesized that the SD model detects signal artifacts caused by the silence at the transition point between two files. This helps the model to segment the audio better. To test this hypothesis we created a similar dataset to the change of emotion experiment but removed silence using a VAD. Results can be seen in~\cref{tab:ns}. We see that our approach achieves better results than SD in both PR-F1 and PC-F1 while having comparable results in R-Val. Although we do see that the model achieves comparable R-Val, we do see a different problem with SD in this setup, \textbf{under-segmentation}. About 18\% of the segments created by the model had only a single segment spanning the entire speech signal and about 50\% had 3 or fewer segments in the speech signal. Our approach had 0\% segmentations with 3 or fewer segments.

\begin{table}[t!]
  \caption{results comparing Speaker Diarization and PMI-A(10) on the change of emotion with no silence experiment\label{tab:ns}}
  \centering
  \begin{tabular}{l|c|c|c}
    \toprule
    \multirow{3}{*}{} & \multicolumn{3}{c}{\textbf{change of emotion no silence}}  \\
    \cline{2-4}
    & \textbf{PR-F1} & \textbf{R-Val} & \textbf{PC-F1}  \\
    \midrule

    SD & 25.3\pmr{0.9}& 42.1\pmr{0.5}& 47.7\pmr{0.8}\\
    PMI & 26.7\pmr{0.5}& 41.5\pmr{0.3}& 66.9\pmr{0.2}\\
    \midrule
  \end{tabular}
  \vspace{-1.5pt}
\end{table}

\section{Ablation}
\label{sec:abl}
Ablation results are presented in Fig.~\ref{figure:abaltion}. We present results using EmoV-DB only, as results for IEMOCAP show similar trends. We report PR-F1 and R-Val only as we observe a high positive correlation (0.55) between PR-F1 and PC-F1.

\newpara{The effect of $k$.} Fig.~\ref{subfigure:C} visualizes the effect of changing the number of segments, $k$, in a constant span selector. Results suggest that increasing the number of segments slightly increases  PR-F1 while decreasing R-Val. This is especially noticeable in emotion change (experiment $1$), where the decrease in R-val is much sharper than the decrease in gender change (experiment $2$). This drop in performance is due to the samples in experiment $1$ usually having fewer segments, which results in a much lower R-Val when increasing $k$.

\newpara{The effect of $v$.} Fig.~\ref{subfigure:v} presents the effect of increasing $v$ on A($v$) span selector. Results suggest that increasing $v$ slightly lowers the PR-F1 while significantly improving R-Val. Using $v=5$ yields a very low R-val. However, when increasing $v$ to $10$, R-Val increases drastically (at a slight cost of PR-F1). A($10$) seems to be the best at balancing between PR-F1 and R-Val.

\newpara{The effect of $t$.} Fig.~\ref{subfigure:T} presents the effect of changing the threshold parameter in the threshold span selector. We observe that increasing the threshold increases PR-F1 in both experiments. The threshold seems to have a different effect on R-Val. Increasing the threshold increases R-val up to some point, in which it starts to have a negative effect. T(-10) seems to be the most balanced threshold.

\newpara{The effect of acoustic-sentence size:} In Fig.~\ref{subfigure:s} we visualize the effect of changing the \emph{acoustic-sentences} sizes. Results suggest that increasing the size of the acoustic-sentences decreases both PR-F1 and R-val, where the size of $0.5$s reaches the best performance.



\section{Discussion}
We presented a new approach for Unsupervised Speech Segmentation and propose a simple and efficient (no training is needed) unsupervised baseline for the approach. We used two data sets and defined four benchmarks that can be used to test and evaluate the performance of different models on this new approach. We empirically show that this pipeline is superior to the evaluated baselines under most metrics while leaving room for improvement in future work. The pipeline also seems to work better for the new approach than pipelines designed for a single acoustic-semantic use case.  Our work presents the first steps in exploring this new approach. We believe and hope this proposed approach will be interesting and valuable to the spoken language modeling community, as it captures higher levels of acoustic-semantic style in speech signals, which may also be unique properties of speech. We believe improvements made on this approach may be beneficial in developing hierarchical speech models, spoken dialogue systems, and spoken language understanding. 

\newpara{Limitations.} The proposed research has three main limitations. (i) Inference time may be relatively long. The proposed pipeline leverages a 350M parameter SLM which requires about 1-2 hours of processing per experiment using a 24GB GPU (A5000). (ii) Although simple and efficient, following the proposed approach to convert raw speech into equally sized acoustic-sentences, is far from ideal and limits the performance of the overall system. (iii) This work defines and sets a benchmark for this approach of Unsupervised Speech Segmentation. However, the datasets we provide are relatively simple and focus only on two acoustic-semantic style changes (emotion and gender). Creating a larger dataset with more complex acoustic-semantic style changes could be an interesting next step that will benefit the community. 

\newpara{Future work.} For future work we plan to mitigate most of the limitations described in the paragraph above. Specifically, we plan to explore variable-length acoustic-sentences by first performing segmentation via common methods, such as~\cite{kreuk2020selfsupervised}. Additionally, we would like to explore and develop more advanced SLMs directly dedicated for this speech segmentation approach, incorporating prosodic features as part of the modeling pipeline (e.g., F0, duration).

\clearpage
\bibliographystyle{IEEEbib}
\bibliography{mybib}

\end{document}